\documentclass[letterpaper, 10 pt, conference]{ieeeconf}  

\IEEEoverridecommandlockouts                              

\usepackage{amsmath} 
\usepackage{amssymb}  
\usepackage{dsfont}
\usepackage{graphicx}

\usepackage{psfrag,graphicx,epsfig}
\usepackage{epstopdf}
\usepackage{xspace}
\usepackage{subfig}
\usepackage{float}
\usepackage{placeins}
\usepackage{multirow}
\usepackage{pgf,tikz}
\usepackage{nowidow}
\usepackage{lineno}
\usepackage{xcolor}
\newcommand{\fig}[1]{Fig.~\ref{#1}}
\newcommand{\tab}[1]{Table~\ref{#1}}
\newcommand{\eq}[1]{(\ref{#1})}
\DeclareMathOperator*{\argmin}{arg\,min}

\usepackage{siunitx}
\usepackage{color}
\usepackage{flushend}
\usepackage{lineno}
\usepackage{gensymb}
\usepackage{algorithmic}
\usepackage{hyperref}
\hypersetup{
    colorlinks=true,
    linkcolor=blue,
    filecolor=magenta,      
    urlcolor=cyan,
    pdftitle={},
    pdfpagemode=FullScreen,
    citecolor=black
    }
\allowdisplaybreaks

\title{\LARGE \bf
Tactile Tool Manipulation
}

\author{Yuki Shirai$^1$, Devesh K. Jha$^2$, Arvind U. Raghunathan$^2$, and Dennis Hong$^1$%
\thanks{$^{\dagger}$ Yuki Shirai and Dennis Hong are with the Department of Mechanical and Aerospace Engineering, University of California, Los Angeles, CA, USA 90095 {\tt\small \{yukishirai4869,dennishong\}@g.ucla.edu}}%
\thanks{$^{\ddagger}$Devesh K. Jha and Arvind U. Raghunathan are with Mitsubishi Electric Research Laboratories (MERL), Cambridge, MA, USA 02139 {\tt\small \{jha,raghunathan\}@merl.com}}}%

\begin{document}
\twocolumn[{%
    \renewcommand\twocolumn[1][]{#1}%
    \maketitle
    \begin{center}
        \centering
        \includegraphics[width=1.0\textwidth]{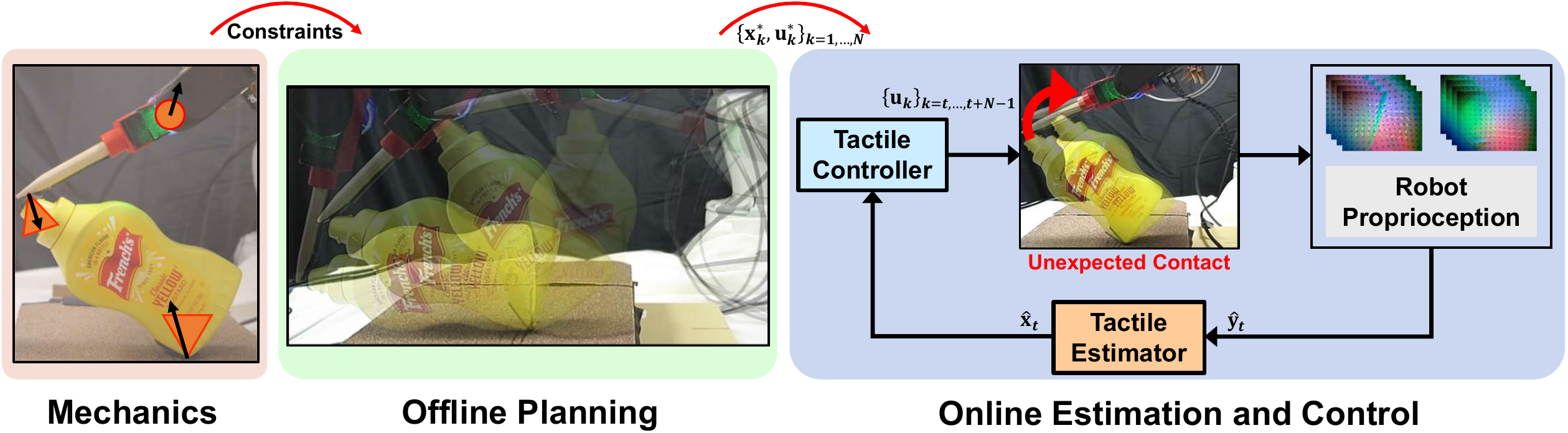} %
    \captionof{figure}{We present tactile tool manipulation where a robot uses an external tool to manipulate an external object. Usage of an external tool results in multiple contact formations which leads to a large number of constraints that need to be satisfied during manipulation. Using the underlying frictional mechanics, we present design of open-loop and closed-loop controllers which can successfully maintain all contact formations during manipulation. We present the design and use of a tactile estimator which makes use of tactile sensing to estimate pose of the system. The tactile estimator is used to perform closed-loop control in an MPC fashion. All hardware experiment videos could be found at \href{https://youtu.be/VsClK04qDhk}{https://youtu.be/VsClK04qDhk}.}
    \label{fig:abst}
    \end{center}%
    }]
    
\footnotetext[1]{Yuki Shirai and Dennis Hong are with the Department of Mechanical and Aerospace Engineering, University of California, Los Angeles, CA, USA 90095 {\tt\small \{yukishirai4869,dennishong\}@g.ucla.edu}}%
 \footnotetext[2]{Devesh K. Jha and Arvind U. Raghunathan are with Mitsubishi Electric Research Laboratories (MERL), Cambridge, MA, USA 02139 {\tt\small \{jha,raghunathan\}@merl.com}}%
\thispagestyle{empty}
\pagestyle{empty}

\begin{abstract}
Humans can effortlessly perform very complex, dexterous manipulation tasks by reacting to sensor observations. In contrast, robots  can not perform reactive manipulation and they mostly operate in open-loop while interacting with their environment. Consequently, the current manipulation algorithms either are inefficient in performance or can only work in highly structured environments. 
In this paper, we present closed-loop control of a complex manipulation task where a robot uses a tool to interact with objects. Manipulation using a tool leads to complex kinematics and contact constraints that need to be satisfied for generating feasible manipulation trajectories. We first present an open-loop controller design using Non-Linear Programming (NLP) that satisfies these constraints. In order to design a closed-loop controller, we present a pose estimator of objects and tools using tactile sensors. Using our tactile estimator, we design a closed-loop controller based on Model Predictive Control (MPC). The proposed algorithm is verified using a 6 DoF manipulator on tasks using a variety of objects and tools. We verify that our closed-loop controller can successfully perform tool manipulation under several unexpected contacts. 



\end{abstract}
\section{Introduction}\label{sec:intro}



Using contacts efficiently can provide additional dexterity to robots while performing complex manipulation tasks~\cite{mason2018toward, doi:10.1126/science.aat8414, shirai_2022iros}. However, most robotic systems avoid making contact with their environment. 
This is mainly because contact interactions lead to complex, discontinuous dynamics and thus, planning, estimation, and control of manipulation require careful treatment of these constraints. As a result of these challenges, most of the classical control approaches are not applicable to control of manipulation systems~\cite{mason2018toward, 8239534, shirai2022chance, 9811812, shirai2023covariance}. However, closed-loop control of manipulation tasks is imperative for design of robust, high-performance robotic systems that can effortlessly interact with their environments. 

In this paper, we consider tool manipulation where 
a robot can grasp an external tool that can be used to pivot an external object in the environment (See \fig{fig:abst}). As could be seen in \fig{fig:not_wrench}, tool manipulation leads to multiple contact formations between the robot \& a tool, the tool \& an object, and the object \& environment. It is easy to imagine that planning for tool manipulation needs to incorporate all constraints imposed by these contact formations. This makes planning for tool manipulation extremely challenging. Furthermore, the robot can not directly observe all the relevant contact and object states during tool manipulation. This imposes additional complexity during controller design. This makes tool manipulation a challenging, albeit extremely rich system to study closed-loop controller design for manipulation. 

We present design of planning, estimation, and control for tool manipulation using tactile sensing. In particular, we first present analysis of the underlying contact mechanics which allows us to plan feasible trajectories for manipulating an external object. To allow robust implementation of the planned manipulation, we design a closed-loop controller using tactile sensors co-located at the fingers of the gripper. We present design of a tactile estimator which estimates the pose of the external object during manipulation. This estimator is used to design a closed-loop controller using MPC. The proposed planner and closed-loop controller are extensively tested with several different tool-object pairs.

\textbf{Contributions.} This paper has the following contributions:
\begin{enumerate}
    \item We present design of closed-loop controller for tool manipulation using tactile sensing and NLP.
    \item The proposed controller is implemented and verified on tasks using different tools and objects using a 6 DoF manipulator equipped with GelSlim tactile sensors.
\end{enumerate}




\section{Related Work}\label{sec:related_works}
Our work is inspired by seminal work on manipulation by shared grasping~\cite{hou2020manipulation} which discusses mechanics of shared grasping and shows impressive demonstrations. 
The task that we present in this paper is a complex version of shared grasping where the robot uses a tool  instead of a rigid end-effector to manipulate objects. This variation leads to additional contact formations. 
These additional constraints make the problem more complicated to plan, control, and estimate compared to those works. 

Model-based planning for tool manipulation was earlier presented in~\cite{holladay2019force}.
Learning-based algorithm of grasping for tool manipulation is presented in \cite{FangIJRR2020}.
In our work, we consider a closed-loop controller and estimator in addition to planning for tool manipulation to robustify the system. 

Our work is also closely related to the remarkable previous work on tactile estimation and reactive manipulation presented in \cite{7989460, 9561781, KETO,  8461175, 9196976, 9811713, 9561646}.
For estimators, \cite{7989460, TerrySeed2022} show a pose estimator for tools  and \cite{9561781} present tactile localization.
Learning-based estimator for tool manipulation using vision is presented in \cite{KETO}. 
In this work, in addition to a tool through tactile sensors, we try to estimate and control a pose of an object, which introduces additional extrinsic contact.
For reactive manipulation, 
our work is closely related to the seminal work presented in~\cite{9196976} where slip detection is used to recompute a new controller that can stabilize the manipulation task. 
\cite{9811713} shows the impressive closed-loop controller by simultaneous design of controller and estimator. 
However, the task in~\cite{9196976} is inherently stable as the object is always grasped by the robot. Also, the tactile sensors can directly estimate the pose of the object, which cannot be done for tool manipulation because tactile sensors are not attached between the object and the end-effector. Compared to  \cite{9811713}, which focuses on regulation of an object  using force / torque sensors, we focus on tracking of tool manipulation using tactile sensors. Furthermore, the current paper considers multiple contact formations which leads to more complex constraints.
%




\section{Mechanics of Tool Manipulation}\label{sec:problem_statement}
In this section, we explain mechanics of tool manipulation as illustrated in \fig{fig:not_wrench} and then discuss Trajectory Optimization (TO) of tool manipulation for designing open-loop trajectories.  Before explaining the details, we present our assumptions in this paper as follows:
\begin{enumerate}
\item The object and the tool are rigid. 
\item The object and the tool always stay in quasi-static equilibrium. 
\item We consider simplified quasi-static mechanics in 2D. 
\item The kinematics of the tool and the friction coefficients for different contact formations are known.
\end{enumerate}
The notation of variables are summarized in \tab{tab:my_label}. We define the rotation matrix from frame $\Sigma_A$ to $\Sigma_B$ as ${ }_{B}^{A} R$. We denote $\mathbf{p}_i$ as a position at contact $i$ defined in $\Sigma_W$. We denote $x$- and $y$-axis as axes in 2D plane and $z$-axis is perpendicular to the plane.

\begin{table}[]
    \centering
        \caption{\textbf{Notation of variables}.  $\Sigma$ column indicates the frame of variables. See \fig{fig:not_wrench} and \fig{tactile_estimator_pipeline} for graphical definition.}
\begin{tabular}{|c|c|c|c|}
\hline Name & Description & Size  &  $\Sigma$ \\
\hline $\mathbf{w}_E$ & reaction wrench at point $A$ & $\mathbb{R}^2$ & $W$ \\
 $\mathbf{w}_O$ & gravity of object at point $O$ & $\mathbb{R}^2$ & $W$ \\
 $\mathbf{w}_{TO}$ & wrench from the tool to the object at point $B$ & $\mathbb{R}^2$ & $T$ \\
  $\mathbf{w}_{T}$ & gravity of tool at point $T$ & $\mathbb{R}^2$ & $W$ \\
  $\mathbf{w}_{G}$ & wrench from the gripper to the tool at patch $C$ & $\mathbb{R}^2$ & $G$ \\
  ${\theta}_{O}$ & orientation of object & $\mathbb{R}^1$ & $W$ \\
    ${\theta}_{T}$ & orientation of tool & $\mathbb{R}^1$ & $W$ \\
      ${\theta}_{G}$ & orientation of gripper & $\mathbb{R}^1$ & $W$ \\
            ${\theta}_{S}$ & relative orientation of frame at center of grasp & $\mathbb{R}^1$ & $S$ \\
\hline 
\end{tabular}
    \label{tab:my_label}
\end{table}

\subsection{Quasi-Static Mechanics of Tool Manipulation}\label{sec:mechanics}
As is shown in \fig{fig:not_wrench}, tool manipulation leads to  several contact formations at $A$, $B$, and $C$ that would need to be maintained during manipulation. Additionally, we need to consider quasi-static equilibrium of the tool and the object in the presence of these contacts.
 The static equilibrium of the  object is described as:
\begin{subequations}
\begin{flalign}
F_O(\mathbf{w}_E, {\mathbf{w}_O},  {{^W_T}R} {\mathbf{w}_{TO}})  = \mathbf{0},\label{Mechanicsforceeq1}\\
G_O(\mathbf{w}_E, {\mathbf{w}_O},  {{^W_T}R} {\mathbf{w}_{TO}}, \mathbf{p}_A, \mathbf{p}_B, \mathbf{p}_O) = 0 \label{Mechanicsmoment_eq1}
\end{flalign}
\label{Mechanics}
\end{subequations}
where $F_O$ and $G_O$ represent static equilibrium of force and moment, respectively. 
%
The static equilibrium of the  tool is:
\begin{subequations}
\begin{flalign}
F_T(\mathbf{w}_T, {^W_G}R{\mathbf{w}_G}, {^W_T}R {\mathbf{w}_{OT}})
 =\mathbf{0},\label{tooleq1}\\
 G_T(\mathbf{w}_T, {^W_G}R{\mathbf{w}_G}, {^W_T}R {\mathbf{w}_{OT}}, \mathbf{p}_B, \mathbf{p}_T, \mathbf{p}_{G1}, \mathbf{p}_{G2}) = 0\label{tooleq2}
\end{flalign}
\label{MechanicsTool}
\end{subequations}
Note that $^T\mathbf{w}_{TO} = -^T\mathbf{w}_{OT}$. In this work, we approximate patch contact at $C$ as two point contacts with the same force distribution, and thus we have $ \mathbf{p}_{C1}, \mathbf{p}_{C2}$ in \eq{tooleq2}. In the next section, we consider contact formations at $A$, $B$, and $C$ while making necessary simplifications for modeling.

\subsection{Contact Model}
\begin{figure}
    \centering
    \includegraphics[width=0.495\textwidth]{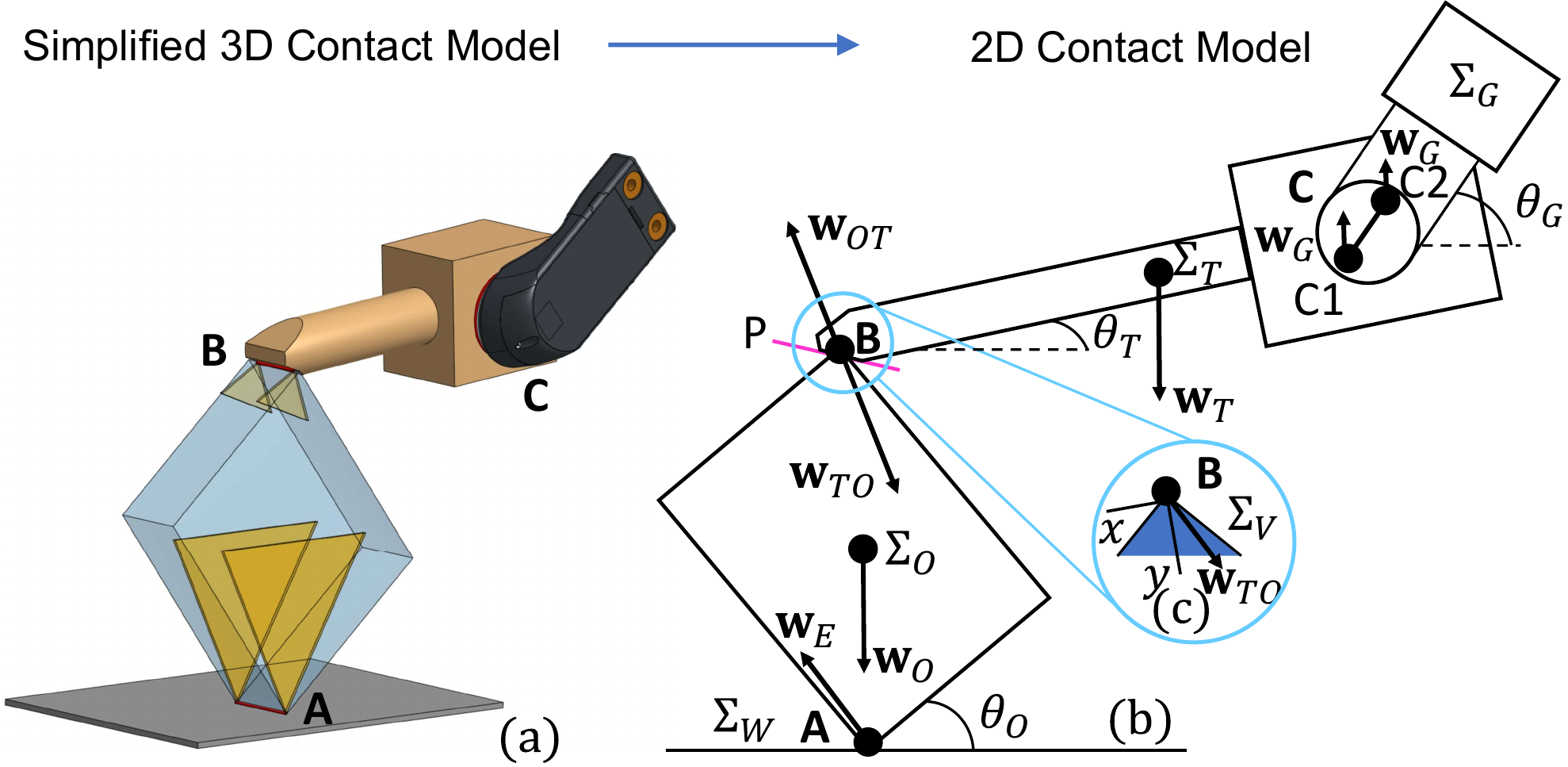}
    \caption{\textbf{Mechanics of tool manipulation.} (a): A simplified 3D contact model for tool manipulation highlighting the three main contact interactions during the task. (b): Free-body diagram of a rigid body and a tool during tool manipulation in 2D. (c): Force from a tool to an object has to lie on a cone defined by the shape of the object.}
    \label{fig:not_wrench}
\end{figure}
We first discuss the contact model in 3D then we present the approximated contact model in 2D using \fig{fig:not_wrench}.
In a simplified 3D setting, the different contact formations could be best described as follows:
\begin{enumerate}
\item contact $A$: line contact.
\item contact $B$: line or patch contact. 
\item contact $C$: patch contact.
\end{enumerate}
For line contacts $A$ and $B$, we need to consider generalized friction cones \cite{erdmann1994representation} to describe sticking line contact in 3D. However, this work considers manipulation in 2D as shown in \fig{fig:not_wrench} (b)  and thus we can argue that there is no moment to break the line contact. Thus, we can approximate line contacts as two point contacts with the same force distribution, leading to the  larger coefficients of friction effectively. 
Also for patch contacts at contact $C$, we need to consider 4D limit surface \cite{xydas1999modeling} where we have 3D force $[f_x, f_y, f_z]$ and 1D moment 
$m_z$. However, in practice, implementing $m_z$ is difficult, especially for position-controlled manipulators with  a force controller with low bandwidth. Thus, this work approximates patch contact at $C$ as two point contacts (see \fig{fig:not_wrench} (b)) with same force distribution. This approximation makes low-level controllers track the  force trajectory easily.


For point contacts $A, B, C_1, C_2$, we have the following friction cone constraints:
\begin{equation}
     -\mu_i f_{y}^i \leq f_{x}^i \leq \mu_i f_{y}^i, f_{y}^i\geq 0,  \forall i = \{A, B, C_1, C_2\}
     \label{FC}
\end{equation}
where $\mu_i$ is the coefficient of friction at contact $i= \{A, B, C_1, C_2\}$ and $f_{x}^i,  f_{y}^i$ are  tangential and normal forces for  each local coordinate. Note that we set $\mu_i = 2\mu_{i, \text{point}}, i= \{A, B\}$ where $\mu_{i, \text{point}}$ is the coefficient of friction between the environment and \textit{point} contact $A, B$ to take into account line contact effects.

\textit{Remark 1}: 
Contact formation at $B$ could be either patch or line contact. To formally discuss the change of these two different contact modes, constraints such as complementarity constraints are required, which is out of scope in this paper. Thus, we assume that contact $B$ always realizes line contact. 

\subsection{Contact between Tool and Object}\label{contact_B_Sec}
The line contact at $B$ introduces an important insight. As illustrated in \fig{fig:not_wrench} (b), this line contact is on a certain plane $P$ created by a tool. The plane $P$ is used to discuss the friction cone between the object and the tool since slipping can only occur along the plane $P$. 
Thus, by changing the orientation of the tool, the orientation of this plane also changes. This does not have an effect on local friction cone constraints \eq{FC} but does have an effect on the object through static equilibrium. 
Furthermore, different tools have different  tip shapes (see \fig{fig:openTO}). Based on kinematics of the tool, local force definition changes, which is tricky and unique to tool manipulation. In conclusion, the system has a preferred orientation of the plane $P$ for finding a feasible trajectory. 


Another unique nature of this task is that we need to explicitly consider the feasible region of a force controller. Note that the manipulator can only apply forces along the axes where its motion is constrained.  This constraint needs to be explicitly enforced during optimization to generate mechanically feasible force trajectories.

Hence, like friction cone constraints, we formulate inequality constraints in vertex frame $\Sigma_V$ (see \fig{fig:not_wrench} (c)) such that $\mathbf{w}_{TO}$ is constrained by the object:
\begin{equation}
     -\rho {f_{y}} \leq f_{x} \leq \rho f_{y}, f_{y}\geq 0
     \label{FC_kin}
\end{equation}
where $[f_{x}, f_{x}]^\top = {^V_T}R\mathbf{w}_{TO}$. We define $\Sigma_V$ such that $y$-axis of $\Sigma_V$ bisect the angle of vertex $B$. $\rho$ can be determined by the shape of the object.

\subsection{Trajectory Optimization for Planning}
We formulate TO for tool manipulation as follows:
\begin{subequations}
\begin{flalign}
\min _{\mathbf{x, u, f}} &\; \sum_{k=1}^{N} (\mathbf{x}_{k} - \mathbf{x}_{g})^{\top} Q (\mathbf{x}_{k} - \mathbf{x}_{g})+\sum_{k=0}^{N-1}\mathbf{u}_{k}^{\top} R \mathbf{u}_{k} \\
\text{s.t.} &\; \eq{Mechanics}, \eq{MechanicsTool}, \eq{FC}, \eq{FC_kin},  \label{to_const1}\\
 &\; \mathbf{x}_{0} = \mathbf{x}_{s}, \mathbf{x}_{N} = \mathbf{x}_{g},
\mathbf{x}_{k} \in \mathcal{X}, \mathbf{u}_{k} \in \mathcal{U}, \mathbf{f}_{k} \in \mathcal{F}\label{bounds_variables}
\end{flalign}
\label{equation_open_TO}
\end{subequations}
where $\mathbf{x}_{k} = [\theta_{O, k}, \theta_{T, k}, \theta_{G, k}]^\top$, $\mathbf{u}_{k} = \mathbf{w}_{G, k}$, $\mathbf{f}_{k} =[\mathbf{w}_{E, k}^\top, \mathbf{w}_{TO, k}^\top]^\top$, $Q=Q^{\top} \geq 0,R=R^{\top} > 0$. $\mathcal{X}$, $\mathcal{U}$, and $\mathcal{F}$ are convex polytopes, consisting of a finite number of linear inequality constraints. 
$\mathbf{p}_i$ can be calculated from  kinematics with $\mathbf{x}_{k}$ since we could assume that contacts ensure sticking contacts by satisfying \eq{FC}. Based on the solution of \eq{equation_open_TO}, we can calculate the pose and force trajectory of the end-effector and we command them during implementation. The resulting optimization in
\eq{equation_open_TO} is NLP, which can be solved using off-the-shelf solvers such as IPOPT \cite{80fe29bf9dc245ffa5c8bd7b3eee2902}.

\textit{Remark 2}: 
For nox-convex shape objects (e.g., peg in \fig{fig:openTO}), the origin of pivoting, $\mathbf{p}_A$, changes over the trajectory. Thus, we cannot directly apply \eq{equation_open_TO} for the non-convex objects. Hence, we solve \eq{equation_open_TO} hierarchically for them where we solve \eq{equation_open_TO} with the first contact origin and then we solve \eq{equation_open_TO} with the next contact origin and so far and so forth.

\section{Tactile Tool Manipulation}\label{sec:tactile_manipulation}
In this section, we present design of our closed-loop controller which makes use of observations from tactile sensors and robot encoders to estimate pose of the system. Most manipulation systems are underactuated and unobservable. The tool manipulation system falls under the same umbrella. Thus, we present the design of a tactile estimator which can estimate $\theta_O$, $\theta_T$, $\mathbf{p}_A$, and the length of the object, $r_O$. Then, we present our MPC-based controller using the estimated states as inputs.


\subsection{Tactile Stiffness Regression}\label{tac_reg_sec}
\begin{figure}
    \centering
    \includegraphics[width=0.495\textwidth]{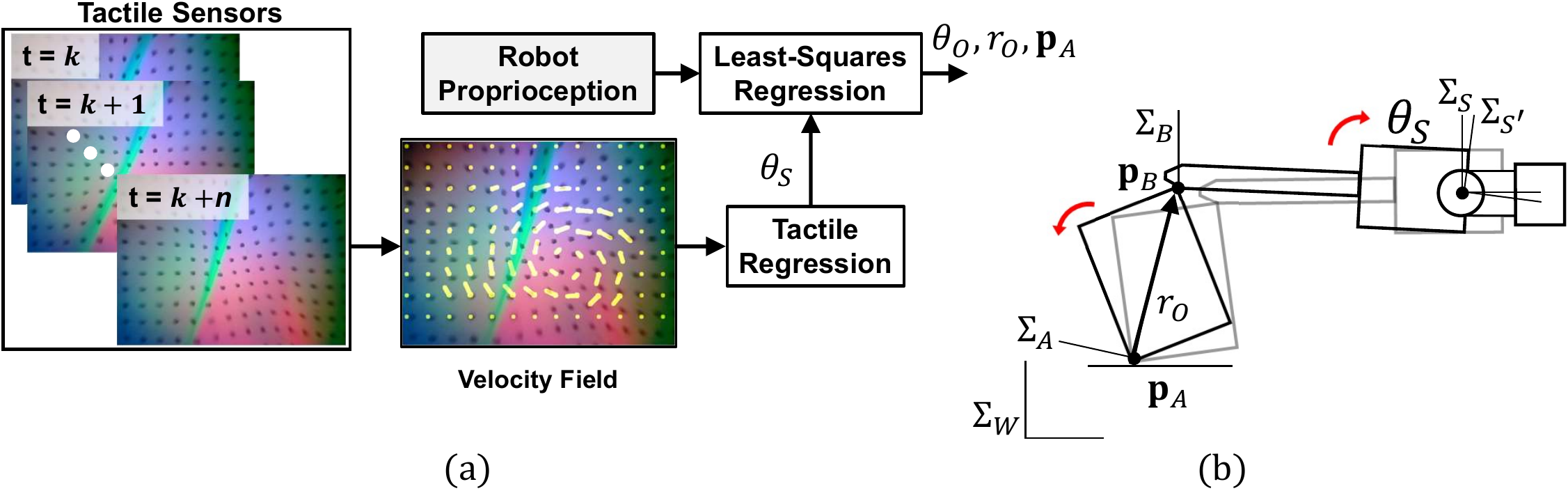} %
    \caption{\textbf{Tactile estimator.}  (a): Given measurements of robot proprioception and tactile sensors, our method estimates the state of the object and the tool. (b): Schematic showing tool manipulation experiencing rotational slipping by $\theta_S$. }
    \label{tactile_estimator_pipeline}
\end{figure}

We use tactile sensors to monitor and estimate the slip of the tool during manipulation. Since the tactile sensors are deformable, we need to identify their stiffness to correctly estimate the slip of objects in grasp. We employ a simple polynomial regression to estimate $\theta_S$ (see \fig{tactile_estimator_pipeline} (b)) given the velocities of all markers as illustrated in \fig{tactile_estimator_pipeline} (a).

We explain how we train the regression model. Given two images at $t = k$ and $t = k + n, n>0$, we compute the velocity flow of the markers on the tactile sensors. 
We use the norm of the velocity flow as input of polynomial regression. We use Apriltag \cite{5979561} to obtain the ground truth values of $\theta_S$ and train the regression algorithm. 
We observe a nonlinear trend in the stiffness of the sensors, i.e., the sensors become more stiff as they deform. 

\subsection{Tactile Estimator}\label{est_tactile}
%
For our estimator design, we make an assumption that contacts at $A$ and $B$ are sticking.
%
%
%
This means that $\mathbf{p}_A$ does not change during pivoting. Using this knowledge, the observed $\mathbf{p}_B$ at $t =k$, denoted as $\bar{\mathbf{p}}_{B, k}$,  can be represented as (see \fig{tactile_estimator_pipeline} (b)):
\begin{equation}
    \left[\bar{\mathbf{p}}_{B, k}^\top, 1\right]^\top = {^W_S}T{^S_{S^{'}}}T(\theta_{S, k}){^{S^{'}}_B}T\left[\mathbf{0}^\top, 1\right]^\top
    \label{kine_chain_1}
\end{equation}
where $\theta_{S, k}$ is the relative rotation of frame at the center of grasp at $t = k$ (we denote this frame as $\Sigma_{S^{'}}$) with respect to the frame at  the reference center of grasp at $t = 0$ (we denote this frame as $\Sigma_{S}$).
${^S_{S^{'}}}T, {^W_{S}}T$ can be obtained from the tactile sensor and encoders, respectively.  ${^{S^{'}}_B}T$ is obtained from the known tool kinematics.
We can represent $\mathbf{p}_B$ at $t=k$, denoted as $\mathbf{p}_{B, k}$, also as follows:
\begin{equation}
    \left[\mathbf{p}_{B, k}^\top, 1\right]^\top = {^W_A}T(\theta_{O, k}, \mathbf{p}_A){^A_B}T(r_O)[\mathbf{0}^\top, 1]^\top
    \label{kine_chain_2}
\end{equation}
%
%
Then,
 using \eq{kine_chain_1} and \eq{kine_chain_2}, with time history of measurements from $ t = 0$ to $ t = m$, 
we can do non-linear regression based on least-squares:
\begin{equation}
    \left\{\theta_{O, k}^*\right\}_{k = 0, \ldots, m}, r_O^*, \mathbf{p}_A^* = \argmin \sum_{k=0}^m\left\|\mathbf{p}_{B, k} - \bar{\mathbf{p}}_{B, k}\right\|^2
    \label{tactile_least_square}
\end{equation}
%
%
Once contact at $\mathbf{p}_A, \mathbf{p}_B$ slip, the estimator is unable to estimate the state of the object anymore like  \cite{9561781}, \cite{9811713}.

\textit{Remark 3}: Similar to \cite{9811713}, our estimator is able to estimate $\theta_{O, k}^*, r_O^*, \mathbf{p}_A^*$. However, similar to~\cite{9811713},  this requires a controller that can maintain the desired contact state during estimation. In this work, we assume that we know the object and tool kinematics during control. Thus, we only make use of $\theta_{O, k}^*$. Controller design when the object kinematics is not known fully is left as a future work.


\textit{Remark 4}: As illustrated in \fig{tactile_estimator_pipeline}, the tool can experience both rotational and translational slip. In practice, we observed that the deformation and high friction at the tactile sensors resulted in minimum translation slip. Thus, we ignored translational slip during manipulation. However, considering the translational slipping might improve the performance of the estimator.

\subsection{Tactile Controller}
Our online controller  based on MPC is as follows:
%
\begin{subequations}
\begin{flalign}
\min _{\mathbf{x, u, f}} &\; \sum_{k=t + 1}^{N +t} ({\theta_{O, k}} - \bar{\theta}_{O, k})^2+\sum_{k=t}^{N + t-1}\mathbf{u}_{k}^{\top} R \mathbf{u}_{k} \\
\text{s.t.} &\; \eq{to_const1}, \eq{bounds_variables}
\end{flalign}
\label{mpc_opt}
\end{subequations}
where $\bar{\theta}_{O, k}$ represent the reference trajectory computed offline using \eq{equation_open_TO}. 
 We observed that slipping between the tool and object kept happening if the controller tracks the tool state as well. Thus, we only consider  state tracking for $\theta_O$ so that the system does not care if $\theta_T$ is tracked - it tries to find a new $\theta_T$ to track $\theta_O$ (i.e., replanning for $\theta_T$). 

\section{Results}\label{sec:results}

\begin{figure*}
    \centering
    \includegraphics[width=0.87\textwidth]{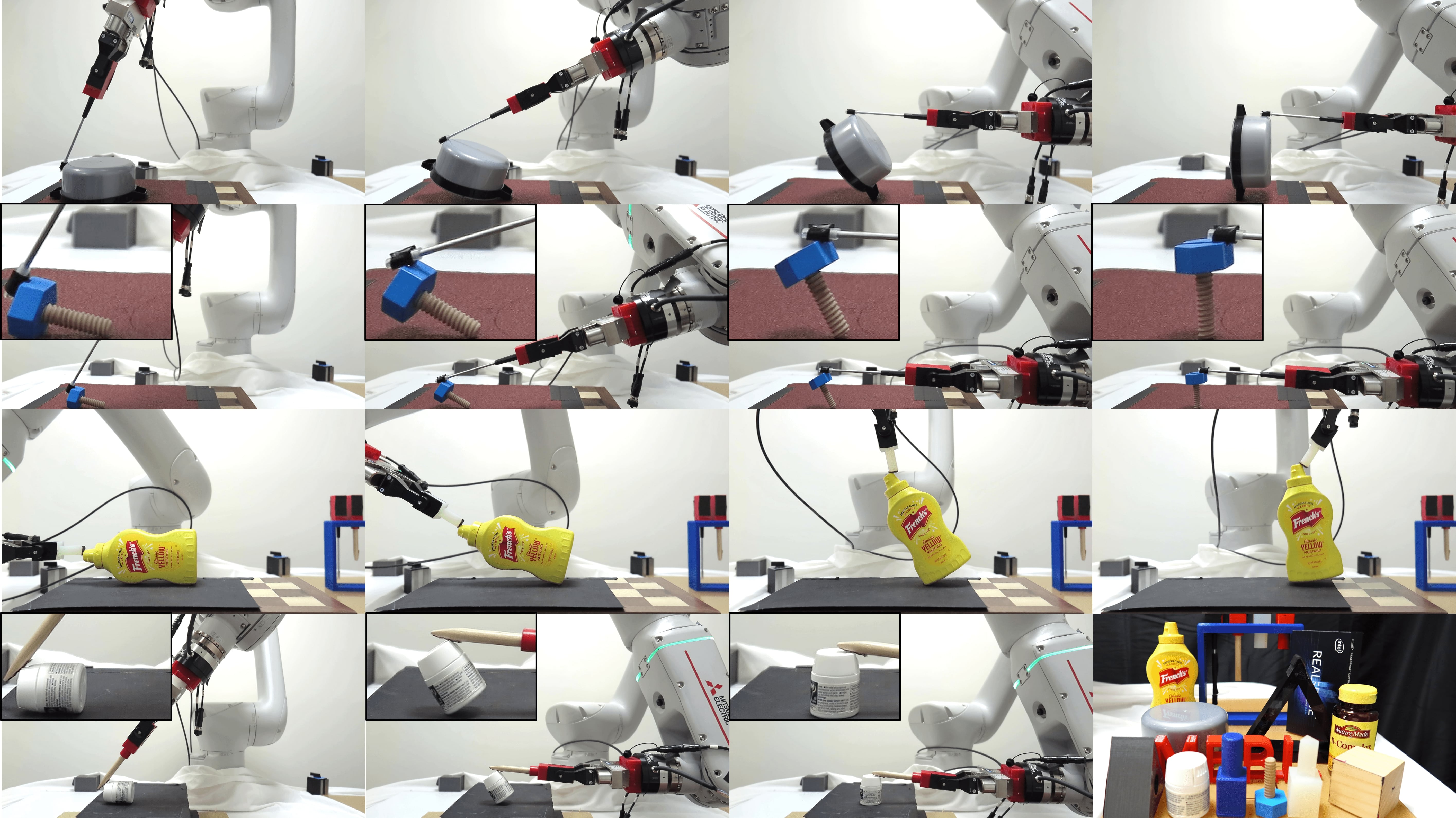} %
    \caption{\textbf{Open-loop tool manipulation.} Our controller could successfully perform tool manipulation with different object-tool-environment pairs. The bottom right picture shows the objects and the tools we use in this paper.}
    \label{fig:openTO}
\end{figure*}

In this section,
we perform several different experiments to answer the following questions:
\begin{enumerate}
\item How do the open-loop trajectories behave on a physical setup?
\item How effective is the proposed closed-loop controller for tool manipulation under different disturbances?
\end{enumerate}

\subsection{Experiment Setup}
For the planner and controller, we use IPOPT \cite{80fe29bf9dc245ffa5c8bd7b3eee2902} with pyrobocop \cite{9812069} interface to solve TO. MPC is run with a large horizon of $N = 160$, and thus can only achieve a control rate of $2$ Hz. However, the control frequency can be increased by using linearized constraints with a QP solver.

%
%
%

For the hardware experiments, we use a Mitsubishi Electric
Assista industrial manipulator arm equipped with a WSG-32 gripper. For the closed-loop experiments, the gripper is equipped with GelSlim 3.0 \cite{9811832} sensors. We use a stiffness controller to track the reference force trajectory \cite{10.1115/1.3140713, jha2022design}. As shown in \fig{fig:openTO}, we test our framework with 12 different objects, 4 different tools, and 5 different environments (i.e., friction surfaces). We use an Apriltag system to obtain the ground truth for pose of objects.



\subsection{Open-Loop Controller}\label{open-loop-results-sec}
In this experiment, we show that our open-loop controller  \eq{equation_open_TO} generates successful trajectories for different objects, tools, and environments.  Note that our framework works even for non-rectangle objects as long as the shape of the object can be approximated as a rectangle in 2D. 
%
The results are shown in \fig{fig:openTO}. More results are shown in the supplementary video. Overall, we verified that our open-loop controller could successfully perform tool manipulation with the \textit{carefully-tuned} parameters by the authors. 

Through these experiments, we observed the following failure cases:
\begin{enumerate}
\item Failure at the beginning of trajectory: We found that the open-loop controller is most susceptible to failure at $t=0$. This is because the tool needs to make contact with the object at $B$. It might happen that the contact force at $B$ is too little that it can not support the moment to lift the object up or in the opposite case, it might be too strong so that the object slips at contact $C$. If too much contact force is applied at contact $B$, the tool might also rotate at the contact $C$.
\item Incorrect physical parameters: We observed that the open-loop controller fails with inaccurate physical parameters such as mass. 
\item Unexpected contacts: Since there is no feedback in the open-loop controller, the manipulation fails if we introduce unexpected contacts during the task.
\end{enumerate}
%
%
Motivated by these failure cases, we discuss how the closed-loop controller can handle them in Sec~\ref{closed_controller_detail_result_sec}. 

\subsection{Tactile Estimator and Controller}\label{closed_discussion_sec}

\subsubsection{Tactile Estimator Results}
\begin{figure}
    \centering
    \includegraphics[width=0.49\textwidth]{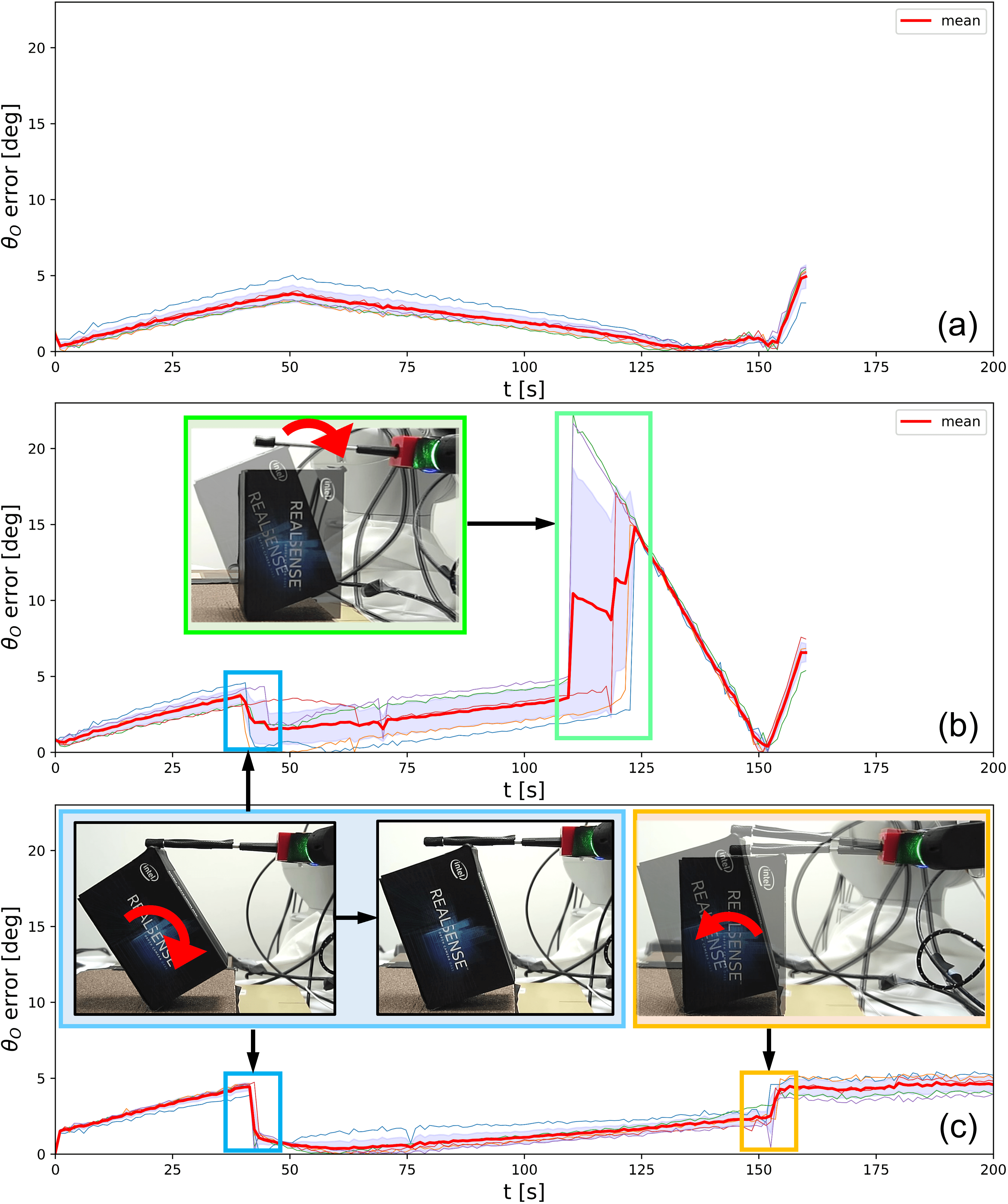} %
    \caption{\textbf{Evaluation of the tactile estimator.} We show the time history of error of $\theta_O$ for $5$ trials (a) with the open-loop controller under no disturbance, (b) with the open-loop controller under disturbance, and (c): with the closed-loop controller under disturbance. The red line shows the mean of and the blue region shows the $95 \%$ confidence interval. We added disturbance around $t = 40$ s for (b) and (c) (see the blue box). For (c), we added another disturbance around $t = 150$ s (see the orange box). The contact is lost around $t = 110$ s for (b) (see the green box). Note that for the open-loop controller, the trajectory runs until $t = 160 $ s because open-loop controller is pre-defined.}
    \label{fig:est_result}
\end{figure}

In this section, we discuss the results of our tactile estimator. To test the accuracy of our estimator, we perform three different kinds of experiments-- the open-loop controller with no external disturbance, the open-loop controller with external disturbance, and the closed-loop controller with external disturbance. In all these experiments, the robot is trying to pivot the same box with the same tool. We perform $5$ trials for each experiment. All results are shown in \fig{fig:est_result}. 

Our estimator works with the  open-loop controller under no disturbances as shown in \fig{fig:est_result} (a)  but does not work under disturbances as shown in  \fig{fig:est_result} (b). Since our estimator assumes that contact is always maintained, once contact is broken (see \fig{fig:est_result} (b) around $t=110$ s), the estimator diverges. In contrast, \fig{fig:est_result} (c) shows that our estimator works  under disturbances since our MPC controller can react to the disturbance and maintain the desired contact state.

\subsubsection{Tactile Controller Results}\label{closed_controller_detail_result_sec}
We demonstrate the effectiveness of our tactile controller to recover from different unexpected contacts in Sec~\ref{open-loop-results-sec}.


We first discuss recovery from slipping of the tool in the gripper fingers, i.e., non-zero $\theta_S$ (see\ \fig{tactile_estimator_pipeline} for definition of $\theta_S$) at $t = 0$ as described in failure case $\# 1$ in Sec~\ref{open-loop-results-sec}.
We implemented the open- and closed-loop controllers with the above disturbance at $t = 0$. We did this experiment for 5 trials per controller. We declare failure if the contact is broken. The result is summarized in \tab{open_closed_comparison_initial_state} (Disturbance 1). For $\theta_S = 5 \degree$, both the open- and the closed-loop controllers could complete the task. However, for $\theta_S = 10 \degree, 15 \degree$, we observed that the open-loop controller lost the contact between the tool and object around $t = 110$ s (see \fig{fig:est_result} (b)) while the closed-loop controller could still successfully conduct the pivoting. 

\begin{figure}
    \centering
    \includegraphics[width=0.475\textwidth]{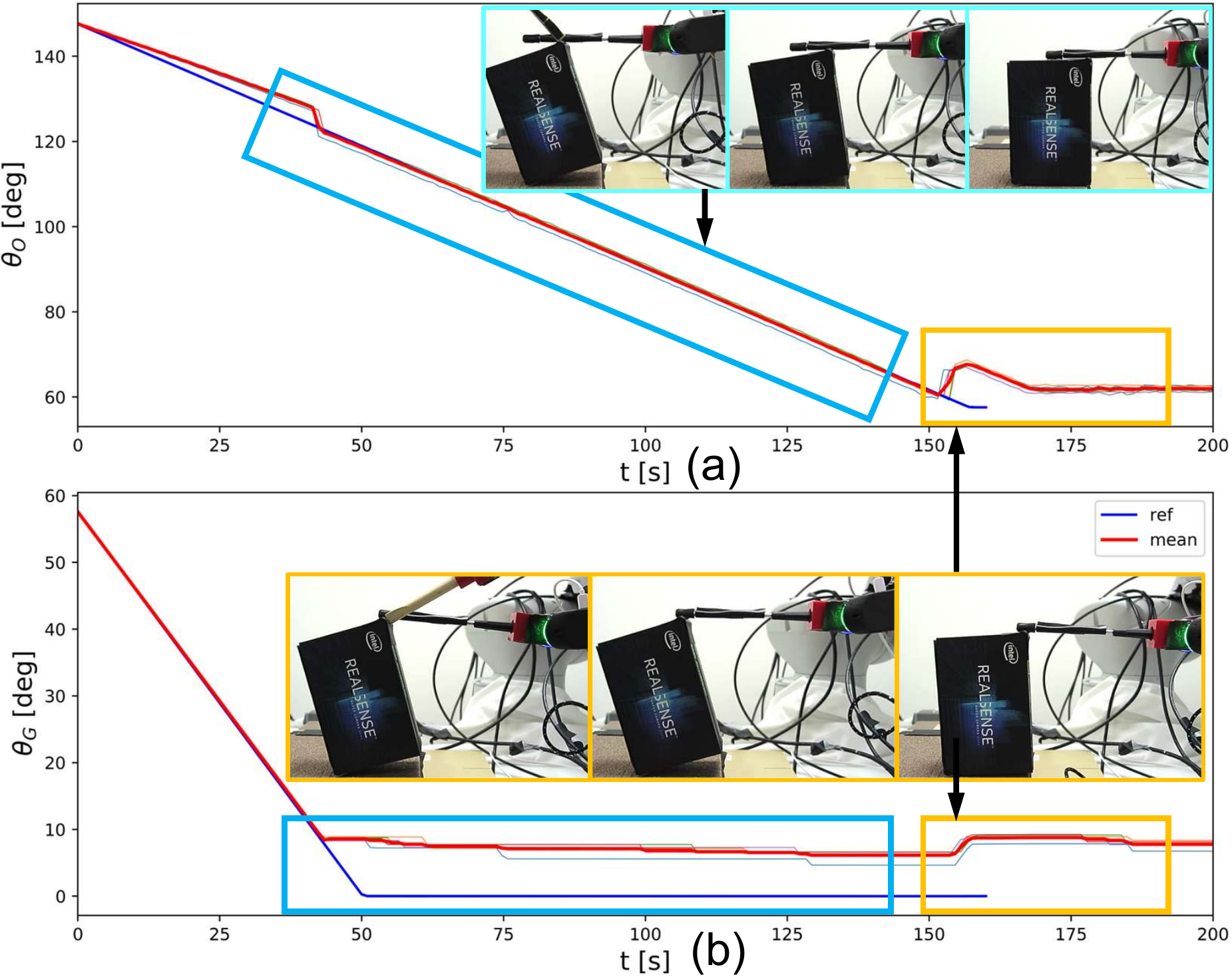} %
    \caption{\textbf{Evaluation of the closed tactile controller.} We show time history of (a) $\theta_O$ and (b) $\theta_G$, with the closed-loop controller under disturbances at $t = 40, 150$ s. The blue line is the reference trajectory computed offline and the red trajectory is the mean of the 5 trajectories computed online.}
    \label{fig:mpc_result}
\end{figure}

Next, we discuss how the closed-loop controller reacts to different unexpected contacts during the trajectory to tackle the failure case $\# 3$ in Sec~\ref{open-loop-results-sec}. 
In these experiments, we add disturbance to the object (see blue and orange box in \fig{fig:est_result} (c)) around $t = 40$ s and $t = 150$ s. We conducted 5 trials. The time history of the object pose $\theta_O$ and the gripper angle $\theta_G$ is shown in \fig{fig:mpc_result}.  
\fig{fig:mpc_result} (a) shows that the closed-controller could successfully track the reference trajectory even under these unexpected contacts. The reactive control efforts can be observed around $t = 40, 150$ s in \fig{fig:mpc_result} (b). The robot changes its gripper orientation to maintain the constraints discussed in Sec~\ref{sec:problem_statement}. The results discussed here are also summarized in \tab{open_closed_comparison_initial_state} (Disturbance 2).

Finally, we demonstrate the closed-loop controller with incorrect mass (failure case $\# 2$ in Sec~\ref{open-loop-results-sec}). In these experiments, we  solve \eq{equation_open_TO} and \eq{mpc_opt} with mass different from the true mass of the object and use the solution for implementation. The results are summarized in \tab{open_closed_comparison_physical_params}. We observed that the closed-loop controller can always successfully pivot the object while the open-loop controller fails especially once $m_O$ is quite different from the true value. The open-loop controller can also work with significantly different $m_O$ as the tactile sensors have significant compliance. This provides some inherent stability to the system during this task. Modeling this compliance and utilizing the model inside MPC as robust tube MPC is an interesting direction \cite{langson2004robust}.




\begin{table}[t]
    \caption{\textbf{Evaluation of the closed tactile controller with disturbances}. The number of successful pivoting attempts of the box over 5 trials for different disturbances are summarized. }
    \centering
\begin{tabular}{|c|c|c|c|c|c|}
\hline \multicolumn{1}{|c}{Box} & \multicolumn{3}{|c|}{ Disturbance 1} & \multicolumn{2}{c|}{ Disturbance 2} \\
\hline  & $5 \degree$ & $ 10 \degree $ & $15 \degree$  & $t = 40$ s & $t = 150$ s \\
\hline Open-loop & $4 / 5$ & $0 / 5$ & $0 / 5$ & $0 / 5$ & N/A \\
\hline Close-loop & $5 / 5$ & $5 / 5$ & $5 / 5$ & $5 / 5$ & $5 / 5$  \\
\hline
\end{tabular}
    \label{open_closed_comparison_initial_state}
\end{table}

\begin{table}[t]
    \caption{\textbf{Evaluation of the closed tactile controller with inaccurate parameters}. The number of successful pivoting attempts of the box over 5 trials for different mass of the object are summarized. The true value of mass of the object is $m_O = 100$ g.}
    \centering
\begin{tabular}{|c|c|c|c|}
\hline $m_O$ [g] & $15$ & $ 200 $ & $1000$   \\
\hline Open-loop & $4 / 5$ & $3 / 5$ & $0 / 5$ \\
\hline Close-loop & $5 / 5$ & $5 / 5$ & $5 / 5$ \\
\hline
\end{tabular}
    \label{open_closed_comparison_physical_params}
\end{table}

\section{Conclusions and Future Work}\label{sec:discussion}
Closed-loop control of manipulation remains elusive. This is because contacts lead to complex, discontinuous constraints that need to be carefully handled. In this paper, we presented tactile tool manipulation. More specifically, we presented the design and implementation of a closed-loop controller to control the complex mechanics of tool manipulation using tactile sensors and NLP. Through extensive experiments, we demonstrate that the proposed method provides robustness against parametric uncertainties as well as unexpected contact events during manipulation. 


There are a number of limitations which we would like to work in the future:


\textbf{Accurate Mechanics of Tool Manipulation}: The natural extension of this work is to consider mechanics in 3D with generalized friction cones \cite{erdmann1994representation}. Additionally, as discussed in Sec~\ref{closed_controller_detail_result_sec}, the system has compliance at the contact locations and we believe that modeling the compliance would lead to a more effective and precise closed-loop controller.

\textbf{Contact-Rich Tool Manipulation}: This work assumes that contact mode (e.g., sticking-slipping, on-off contact) does not change over the trajectory. Thus, our method works with relatively high friction coefficients. However, during the  experiments, we observed that the tool and the object can slip and lose contact, which makes the estimator and the controller diverge. Thus, it would be a promising direction if the extended framework can consider hybrid dynamics which can allow the system to change modes during operation.

\textbf{Analysis of Controllability and Observability for Dexterous Manipulation}: 
One of the fundamental questions that remains open in manipulation is that of controllability and observability.
There have been remarkable works in controllability and observability for manipulation \cite{doi:10.1177/027836499901800105, 897777, 660870}. However, the theory of  controllability and observability is limited to more dexterous manipulation (e.g., tool manipulation). This limits the generality of model-based controller design for manipulation. Therefore, it would be useful to understand and study controllability and observability for frictional interaction tasks. 









\bibliographystyle{IEEEtran}
\bibliography{main}

\end{document}